%% file: paper.tex
\documentclass[10pt,twocolumn,letterpaper]{article}

\usepackage{subfig}
\usepackage{cvpr}
\usepackage{graphicx}
\usepackage{amsmath}
\usepackage{amssymb}
\usepackage{booktabs}
\usepackage{bm}
\usepackage{multirow}
\usepackage{makecell}
\usepackage{stfloats}
\usepackage{bbding}

\usepackage{color}
\input{vcl-shortcuts.tex}
\usepackage[pagebackref=true,breaklinks=true,letterpaper=true,colorlinks,bookmarks=false]{hyperref}

\ifcvprfinal\fi

\def\cvprPaperID{6402} 
\def\confName{CVPR}
\def\confYear{2022}

\begin{document}

\title{Semi-supervised Implicit Scene Completion from Sparse LiDAR}

\author{Pengfei Li$^{1}$, Yongliang Shi$^{1}$, Tianyu Liu$^{2}$, Hao Zhao$^{3}$, Guyue Zhou$^{1}$ and Ya-Qin Zhang$^{1}$}

\makeatletter
\g@addto@macro\@maketitle{
  \begin{figure}[H]
  \setlength{\linewidth}{\textwidth}
  \setlength{\hsize}{\textwidth}
  \vspace{-7mm}
  \centering
 	\begin{tabular}{@{}c@{\hspace{1mm}}c@{}}
 	\includegraphics[width=1\textwidth]{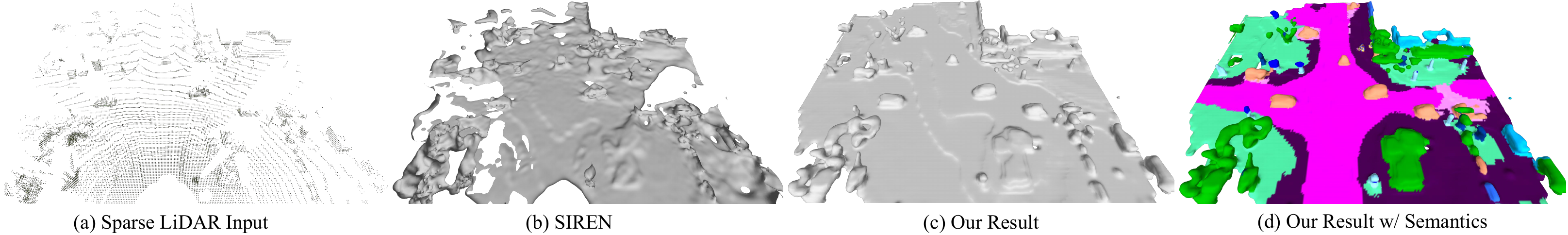}
 	\end{tabular}
  \vspace{2mm}
  \caption{(a) The input is a sparsity-variant point cloud of road scenes captured by LiDAR. (b) The implicit fitting result of a recent method named SIREN \cite{c10}. Note that this is well tuned by an exhaustive parameter search. (c) The output of our method is a neural signed distance function of arbitrary resolution, i.e., implicit scene completion. (d) Our result with optional semantic parsing.}
  \label{fig:teaser}
  \vspace{2mm}
  \end{figure}
}
\makeatother

\maketitle
\begin{abstract}
Recent advances show that semi-supervised implicit representation learning can be achieved through physical constraints like Eikonal equations. However, this scheme has not yet been successfully used for LiDAR point cloud data, due to its spatially varying sparsity. In this paper, we develop a novel formulation that conditions the semi-supervised implicit function on localized shape embeddings. It exploits the strong representation learning power of sparse convolutional networks to generate shape-aware dense feature volumes, while still allows semi-supervised signed distance function learning without knowing its exact values at free space. With extensive quantitative and qualitative results, we demonstrate intrinsic properties of this new learning system and its usefulness in real-world road scenes. Notably, we improve IoU from 26.3\% to 51.0\% on SemanticKITTI. Moreover, we explore two paradigms to integrate semantic label predictions, achieving implicit semantic completion. Code and models can be accessed at \url{https://github.com/OPEN-AIR-SUN/SISC}.
\end{abstract}

\footnotetext[1]{Institute for AI Industry Research (AIR), Tsinghua University, China
        lipengfei181@mails.ucas.ac.cn, ylshi@bit.edu.cn, \{zhouguyue, zhangyaqin\}@air.tsinghua.edu.cn}
\footnotetext[2]{The Hong Kong University of Science and Technology, China    
        tianyu.liu@connect.ust.hk}
\footnotetext[3]{Intel Labs China, Peking University, China  
        zhao-hao@pku.edu.cn, hao.zhao@intel.com}

\section{Introduction}
\label{sec:introduction}

\input{tex/introduction.tex}

\section{Related Work}
\label{sec:related}

\input{tex/related-work.tex}

\section{Formulation}
\label{sec:formulation}
\input{tex/formulation.tex}

\section{Method}
\label{sec:method}

\input{tex/method.tex}




\section{Experiments}
\label{sec:experiments}
\input{tex/experiments.tex}

\section{Conclusion}
\label{sec:conclusion}
\input{tex/conclusion.tex}


{\small
	\bibliographystyle{ieee}
	\bibliography{paper}
}

\clearpage

\section{Supplementary Material}
\label{sec:appendix}
\input{tex/supplement.tex}

\end{document}

%% file: vcl-shortcuts.tex
\usepackage{times}
\usepackage{epsfig}
\usepackage{graphicx}
\usepackage{float}
\usepackage{wrapfig}
\usepackage{amsmath,amssymb,amsthm}
\usepackage{algorithm,algorithmicx,algpseudocode}
\usepackage{bm,xspace}
\usepackage{comment}
\usepackage{verbatim}
\usepackage{multirow}
\usepackage{balance}
\usepackage{url}
\usepackage{booktabs}
\usepackage{etoolbox,siunitx}
\usepackage{calc}
\usepackage{pifont,hologo}

\newcommand{\ra}[1]{\renewcommand{\arraystretch}{#1}}

\usepackage[super]{nth}
\usepackage{nicefrac}
\robustify\bfseries

\DeclareMathSymbol{@}{\mathord}{letters}{"3B}

\def\latex/{\LaTeX}
\def\bibtex/{\hologo{BibTeX}}


%% file: tex/introduction.tex
Representing 3D data with neural implicit functions is actively explored recently due to its strong modeling capability and memory efficiency. While most methods are fully supervised \cite{c17}\cite{c18}\cite{c19}, a recent one named SIREN \cite{c10} proposes a semi-supervised formulation to represent shapes with signed distance functions (SDFs). Instead of using expensive ground truth SDF values everywhere, it does not require SDF values at free space thanks to the usage of an Eikonal equation. 

However, even after an exhaustive parameter search, SIREN fails to fit sparse LiDAR data (Fig.~\ref{fig:teaser}-a), which limits its application in many important scenarios such as autonomous driving. This is understandable as SIREN is a pure generative model and LiDAR point clouds are extremely sparse. Specifically, reasons are three-fold: 1) The sparsity of on-surface points amplifies the negative impact of wrongly sampled off-surface anchors. 2) The normal orientations of sparse points cannot be estimated accurately from their neighbors, which serve as a necessary boundary value constraint for SIREN fitting. 3) Without trustworthy boundary values, enforcing a hard Eikonal constraint leads to even inaccurate SDF values. As shown in Fig.~\ref{fig:teaser}-b, the SIREN fitting result is fragmented.

To overcome these limitations, we develop a novel semi-supervised implicit formulation by introducing an intermediate shape embedding domain. Instead of directly fitting a function to map 3D Cartesian coordinates to signed distance values, we first map the Euclidean space to a corresponding high-dimensional shape embedding space, and then the signed distance space. These shape embeddings function as dense boundary values that entangles both zeroth-order (on-surface points) and first-order (normal directions) constraints, in a data driven manner. Naturally, the issue of enforcing a hard Eikonal constraint is also alleviated. Our result is significantly better than SIREN (Fig.~\ref{fig:teaser}-c).

Specifically, we propose a novel hybrid architecture combining a discriminative model and a generative model, along with a tailored training paradigm. In contrast to former arts that turn sparse input into dense output, the proposed paradigm treats both raw sparse data and ground truth dense data as inputs. The discriminative part of our method exploits the strong representation learning power of sparse convolution, generating latent shape codes from sparse point cloud input. The generative model takes as input the ground truth point cloud coordinates along with point-wise latent shape codes retrieved by trilinear sampling, and predicts SDF values of these points. 

Furthermore, we extend our method in two ways for implicit semantic completion: 1) adding a dense discriminative head to predict semantic completion results which can be mapped to the implicit function using K-Nearest-Neighbors; 2) adding a parallel implicit generative head to directly model the implicit semantic label field of the scene. We evaluate on the public benchmark SemanticKITTI, and achieve robust semantic completion results (Fig.~\ref{fig:teaser}-d).

To summarize, our contributions are as follows:

\begin{itemize}
\item[$\bullet$] We develop a semi-supervised implicit representation formulation that incorporates learned shape embeddings as dense boundary value constraints. 
\item[$\bullet$] We apply the proposed formulation in road scene understanding, leading to the first semi-supervised implicit road scene completion method.
\item[$\bullet$] We expand our method with a semantic module and evaluate on the public benchmark SemanticKITTI, achieving good quantitative and qualitative results. Code, data and models are released.
\end{itemize}

%% file: tex/related-work.tex
The general principle of neural implicit representation is to train a neural network to approximate a continuous function that is hard to parameterize otherwise. \cite{c17} proposes to learn deep signed distance functions conditioned on shape codes. Online optimization of the codes leads to impressive completion results. \cite{c18} approximates occupancy functions with conditional batchnorm networks. \cite{c13} introduces data-driven shape embeddings into occupancy networks for indoor scene completion. \cite{c19} uses hyperplanes as compact implicit representations to reconstruct shapes sharply and compactly. \cite{c10} shows that using gradient supervision allows semi-supervised SDF learning and sine activation functions are critical to its success. \cite{c29} combines Gaussian ellipsoids and implicit residuals to represent shapes accurately. These are some recent works that exploit 3D implicit representations for instance-level understanding from point cloud \cite{c30} or RGB \cite{c22} inputs. 

2D implicit representations \cite{c23} have been shown effective for super-resolution. Neural radiance fields \cite{c14} have revolutionized view synthesis. \cite{c24}\cite{c25} achieve real-time radiance field rendering through predicting the coefficients of spherical harmonics. \cite{c26} proposes continuous Multiplane images that allows natural conditioning on single-view inputs. \cite{c27} shows that sparse semantic labels can be effectively propagated when treated in an implicit way. \cite{c28} proposes a promising framework that addresses the composition of radiance fields.

Entangling semantic understanding and scene completion is an intriguing idea as these two sub-tasks may enhance each other. Older works \cite{c15}\cite{c16} have explored the possibility of building a joint random field formulation with both semantic and geometric cues. Room layout estimation \cite{c20}\cite{c21}\cite{c31}\cite{c32} is another old semantic reconstruction formulation. After the advent of deep learning, semantic scene completion \cite{c1} appears as a natural and elegant formulation.  \cite{c3} introduces sparse convolution into the task and identifies the fact that using spatial groups along with sparse convolution
can linearly reduce computation. \cite{c4} shows the effectivenes of incorporating RGB inputs. Its framework is developed in \cite{c7}, via using a conditional variational encoder to capture the distribution of 3D sketches. \cite{c5} demonstrates the importance of weight balancing. \cite{c6} focuses on domain-adaptive semantic understanding with completion. \cite{c8}\cite{c9} are state-of-the-art semantic road scene completion architectures, highlighting point-voxel interaction and lightweight design respectively.



%% file: tex/formulation.tex
\begin{figure*}[t]
\centerline{\includegraphics[width=1\textwidth]{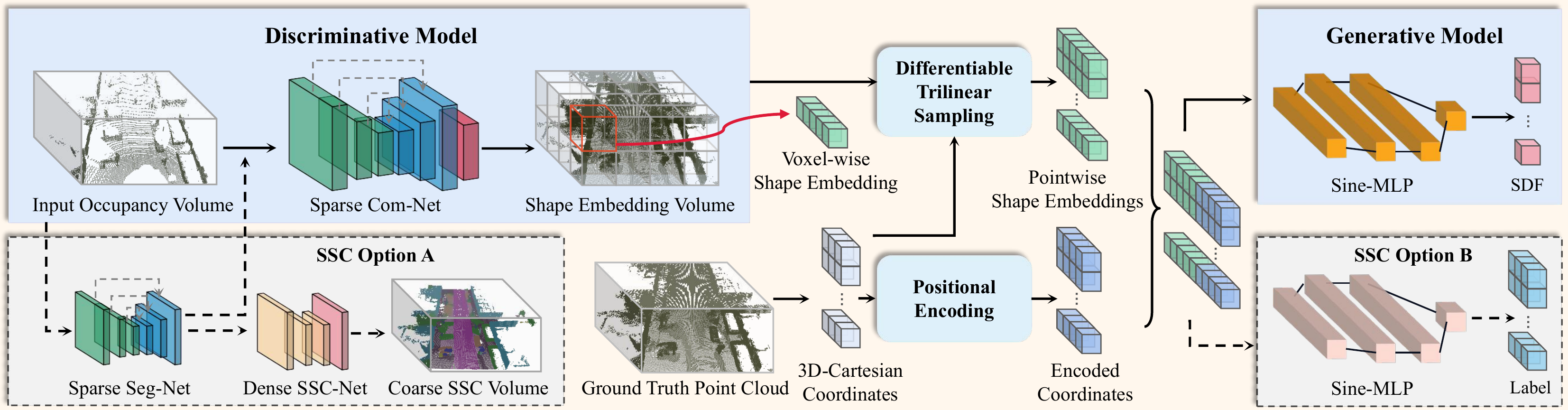}}
\vspace{2mm}
\caption{\textbf{Overview of our architecture.} The discriminative model and generative model is bridged by a differentiable triliner sampling layer. Two options for implicit semantic scene completion (SSC) are outlined in dashed boxes.}
\label{fig:mainfig}
\end{figure*}

The signed distance is the nearest distance from a point of interest to the scene surface, with the sign denoting whether the point is located outside (positive) or inside (negative) of the surface.  And the iso-surface where the signed distance equals zero implicitly delineates the scene. Formally, our goal is to find a function $\Phi(\textbf{x})$ to approximate the underlying signed distance function (SDF), which satisfies a set of $M$ constraints $\mathcal{C}_m$. Each constraint relates the function $\Phi(\textbf{x})$ or its gradient to certain input quantities
$\textbf{a}(\textbf{x})$ on their corresponding domain $\Omega_m$:
\begin{equation}
\begin{split}
\mathcal{C}_m (\textbf{a}(\textbf{x}),&\Phi(\textbf{x}),\nabla_\textbf{x}\Phi(\textbf{x})) = 0, \\
&\forall \textbf{x} \in \Omega_m, m=0,...,M-1.
\end{split}
\end{equation}
Specifically, these constraints are required:
\begin{equation}
\mathcal{C}_0 := |\nabla_\textbf{x}\Phi(\textbf{x})| - 1, \textbf{x} \in \Omega_0.
\end{equation}
\begin{equation}
\mathcal{C}_1 := \nabla_\textbf{x}\Phi(\textbf{x}) - \textbf n(\textbf x), \textbf{x} \in \Omega_1.
\end{equation}
\begin{equation}
\mathcal{C}_2 := \Phi(\textbf{x}) - \rm{SDF}(\textbf{x}), \textbf{x} \in \Omega_2.
\end{equation}
Here, $\mathcal{C}_0$ guarantees $\Phi(\textbf{x})$ satisfies the Eikonal equation in the whole physical space of interest $\Omega_0$, which is a intrinsic property of SDF. $\mathcal{C}_1$ forces that the gradients of $\Phi(\textbf{x})$ equal the normal vectors for input on-surface points in $\Omega_1$. $\mathcal{C}_2$ constrains the values of $\Phi(\textbf{x})$ equal the ground truth SDF for anchor points in $\Omega_2$. In this way, the problem can be regarded as a Eikonal boundary value problem, where the differential equation $\mathcal{C}_0$ is solved under the first-order constraint $\mathcal{C}_1$ and the zeroth-order constraint $\mathcal{C}_2$.

However, the ground truth SDF values at free space are difficult to obtain. A recent method named SIREN \cite{c10} proposes an intriguing variant where the domain of $\mathcal{C}_2$ is limited to on-surface points in $\Omega_1$. As the ground truth SDF values of points in $\Omega_1$ are zero, $\mathcal{C}_2$ is reduced to:
\begin{equation}
\mathcal{C}_2 := \Phi(\textbf{x}), \textbf{x} \in \Omega_1.
\end{equation}
To remedy the lack of constraints on off-surface points, SIREN introduces another constraint:
\begin{equation}
\mathcal{C}_3 := \psi(\Phi(\textbf{x})), \textbf{x} \in \Omega_3.
\end{equation}
Here, $\psi$ pushes $\Phi(\textbf{x})$ values away from 0, for randomly and uniformly sampled off-surface points in $\Omega_3 \subseteq \Omega_0\setminus\Omega_1$.

Nevertheless, this set of constraints fails to address the scenario where on-surface points in $\Omega_1$ are sampled from sparse LiDAR data. Reasons are three-fold: 1) The sparsity of on-surface points in $\Omega_1$ amplifies the negative impact of $\mathcal{C}_3$ on the wrongly sampled off-surface anchors in $\Omega_3$ (i.e., located on or near the surface). 2) The normal orientations of sparse points in $\Omega_1$ cannot be estimated accurately from their neighbors, leading to an incorrect constraint $\mathcal{C}_1$. 3) Without trustworthy boundary value constraints $\mathcal{C}_3$ and $\mathcal{C}_1$, enforcing the hard Eikonal constraint $\mathcal{C}_0$ leads to even inaccurate SDF values at free space.

To overcome these limitations, we propose a novel formulation $\Phi(\textbf{x},\textbf{e})|_{\textbf{e}=\zeta(\textbf{x}, \Omega_1)}$ to approximate SDF. Here, we use $\zeta(\cdot, \cdot)$ to first map the Euclidean space to a high-dimensional shape embedding space. It functions as a dense boundary value constraint for the differential equation. Then $\Phi(\cdot, \cdot)$ maps the shape embedding space to the signed distance space. As a result, the constraints needed to be satisfied are formally re-written as:
\begin{equation}
\mathcal{C}_0^\prime := |\nabla_\textbf{x}\Phi(\textbf{x},\textbf{e})|_{\textbf{e}=\zeta(\textbf{x}, \Omega_1)}| - 1, \textbf{x} \in \Omega_0.
\end{equation}
\begin{equation}
\mathcal{C}_4 := \rho(\zeta(\textbf{x}, \Omega_1)), \textbf{x} \in \Omega_0.
\end{equation}
We use $\rho(\zeta(\textbf{x}, \Omega_1))$ to represent the underlying dense constraint contained in the shape embedding space, which implicitly entangles correct $\mathcal{C}_1$, $\mathcal{C}_2$ and $\mathcal{C}_3$ constraints of the modified formulations:
\begin{equation}
\mathcal{C}_1^\prime := \nabla_\textbf{x}\Phi(\textbf{x},\textbf{e})|_{\textbf{e}=\zeta(\textbf{x}, \Omega_1)} - \textbf n(\textbf x), \textbf{x} \in \Omega_1^\prime.
\end{equation}
\begin{equation}
\mathcal{C}_2^\prime := \Phi(\textbf{x},\textbf{e})|_{\textbf{e}=\zeta(\textbf{x}, \Omega_1)}, \textbf{x} \in \Omega_1^\prime.
\end{equation}
\begin{equation}
\mathcal{C}_3^\prime := \psi(\Phi(\textbf{x},\textbf{e})|_{\textbf{e}=\zeta(\textbf{x}, \Omega_1)}), \textbf{x} \in \Omega_3^\prime.
\end{equation}
where $\Omega_1^\prime$ contains the dense ground truth on-surface points and $\Omega_3^\prime \subseteq \Omega_0\setminus\Omega_1^\prime$. Hence the aforementioned problem of trustworthy boundary values is resolved. Naturally, the issue of enforcing a hard Eikonal constraint is also alleviated.

We implement our representation in a data driven manner. The acquisition of functions $\zeta(\cdot, \cdot)$ and $\Phi(\cdot,\cdot)$ can be cast in a loss function that penalizes deviations from the constraints $\mathcal{C}_0^\prime$, $\mathcal{C}_1^\prime$, $\mathcal{C}_2^\prime$ and $\mathcal{C}_3^\prime$ on
their domain:
\begin{equation}
\begin{split}
\mathcal{L}_{\rm{SDF}} 
&= \lambda_1\int_{\Omega_0}  \left\| |\nabla_{\textbf{x}} \Phi(\textbf{x},\textbf{e})|_{\textbf{e}=\zeta(\textbf{x}, \Omega_1)}| -1 \right\|d{\textbf{x}} \\
&+ \lambda_2\int_{\Omega_1^\prime} (1-\langle \nabla_{\textbf{x}} \Phi(\textbf{x},\textbf{e})|_{\textbf{e}=\zeta(\textbf{x}, \Omega_1)}, \textbf{n}(\textbf{x}) \rangle)d{\textbf{x}} \\
&+ \lambda_3\int_{\Omega_1^\prime} \left\| \Phi(\textbf{x},\textbf{e})|_{\textbf{e}=\zeta(\textbf{x}, \Omega_1)}\right\|d{\textbf{x}} \\
&+ \lambda_4\int_{\Omega_3^\prime} \psi(\Phi(\textbf{x},\textbf{e})|_{\textbf{e}=\zeta(\textbf{x}, \Omega_1)})d{\textbf{x}}.
\label{equ:sdfloss}
\end{split}
\end{equation}
where $\lambda_1$ - $\lambda_4$ are constant weight parameters, $\langle \cdot,\cdot \rangle$ calculates cosine similarity. 

Specifically, we propose a hybrid neural network architecture combining a discriminative model with a generative model, as shown in Fig.~\ref{fig:mainfig}. The discriminative part of our method exploits the strong representation learning power of sparse convolution, generating latent shape embeddings from sparse input $\Omega_1$. It together with the differentiable trilinear sampling module works as function $\zeta(\cdot, \cdot)$. The generative model takes as input the ground truth points $\Omega_1^\prime$ along with point-wise latent shape codes, and predicts SDF values of these points. It functions as $\Phi(\cdot,\cdot)$. Using gradient descent, we can get the optimized $\zeta(\cdot, \cdot)$ and $\Phi(\cdot,\cdot)$ in the parameterized form.

%% file: tex/method.tex

\subsection{Discriminative Model}
\label{subsection:D_Model}
Intuitively, the outdoor scene has the characteristic of repetition. Therefore, convolutional neural network can be employed to exploit the translation-invariance. To this end, taking LiDAR points $\Omega_1$ as input, we firstly conducts voxelization $f_{\rm{vox}}$ to obtain 3D occupancy volume $V_{\rm{occ}}$ with size $1 \times D_{\rm{occ}} \times W_{\rm{occ}} \times H_{\rm{occ}}$. Then the convolutional discriminative model $f_{\rm{dis}}$ maps it into a shape embedding volume $V_{\rm{se}}$ with size $d_{\rm{se}} \times D_{\rm{se}} \times W_{\rm{se}} \times H_{\rm{se}}$:
\begin{equation}
f_{\rm{vox}}(\Omega_1) = V_{\rm{occ}}, f_{\rm{dis}}(V_{\rm{occ}}) = V_{\rm{se}}.
\end{equation}
Here, $d_{\rm{se}}$ is the dimension of the shape embedding outputs.

\begin{figure}[tpb]
\centerline{\includegraphics[width=0.45\textwidth]{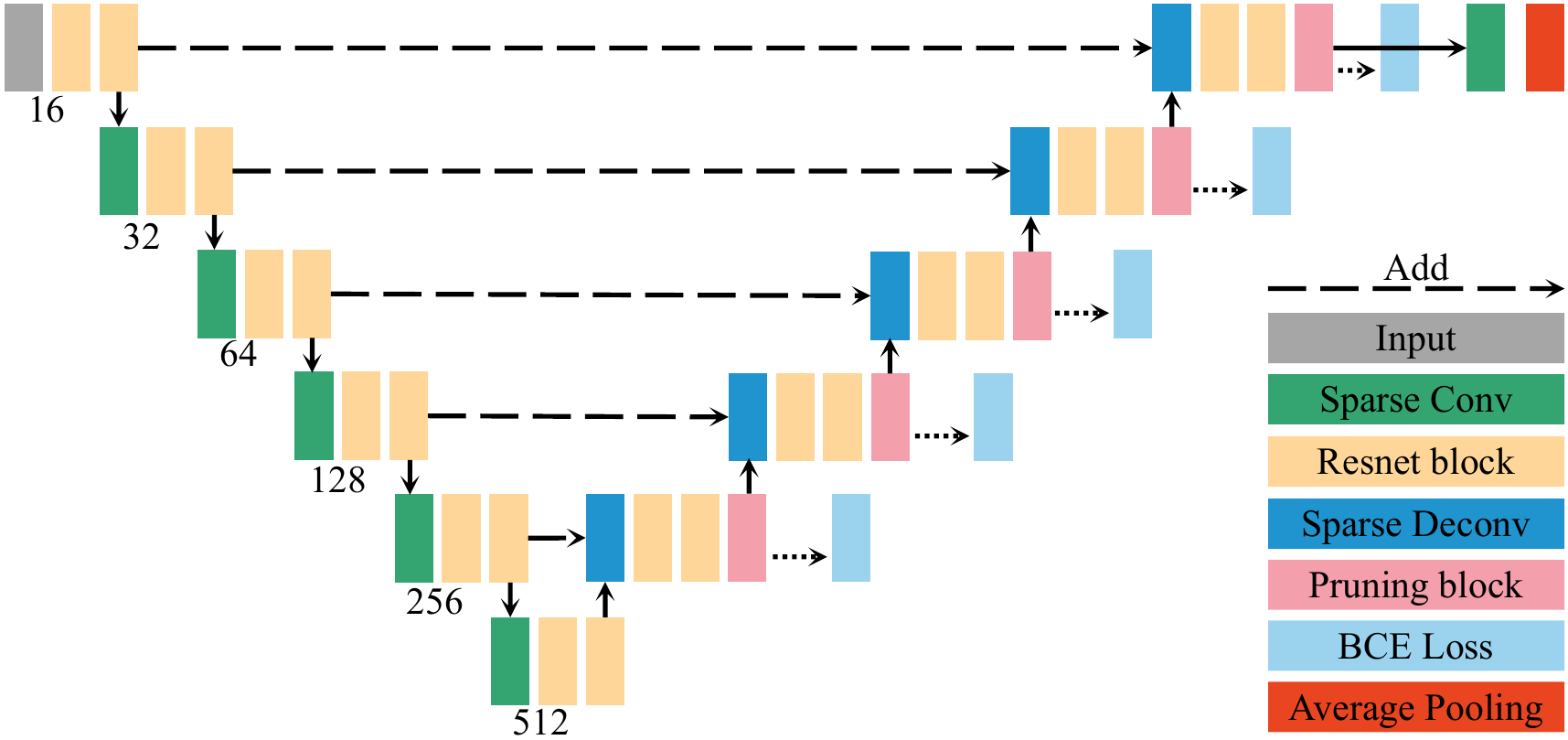}}
\caption{Network structure of Com-Net.}
\label{fig:unet}
\end{figure}

To tackle the sparsity of the input occupancy volume $V_{\rm{occ}}$, we employ the Minkowski Engine \cite{c12} to build our model. Specifically, we use a multiscale encoder-decoder architecutre \cite{c33} named Com-Net as our discriminative model (Fig.~\ref{fig:unet}). It predicts shape embeddings via a shape completion process. The encoder modules consist of convolutional blocks and residual blocks. The decoder modules involve generative deconvolutional blocks to generate new voxels. Yet the constant generation of new voxels will destroy the sparsity just as the \emph{submanifold dilation problem} \cite{c11}. To avoid this, we use a pruning block to prune off redundant voxels. It contains a convolutional layer to determine the binary classification results of whether a voxel should be pruned.  We supervise the classification results of the pruning blocks with binary cross entropy loss:
\begin{equation}
\begin{split}
\mathcal{L}_{\rm{com}} = -{\frac 1 m}\sum_{i=1}^m & {\frac 1 {n_i}} \sum_{j=1} ^ {n_i} [y_{i,j} {\rm{log}}(p_{i,j}) \\
&+ (1-y_{i,j}){\rm{log}}(1-p_{i,j})].
\end{split}
\end{equation}
where $m$ is the count of supervised blocks, $n_i$ denotes the count of voxels in the $i$-th block, $y_{i,j}$ and $p_{i,j}$ are the true and predicted existence probability for voxel $i$ respectively.

In this way, the Com-Net divides the whole scene into several cubes with edges of length $b$, aggregating the latent shape information of $b^3$ voxels into a single one. Thus we obtain the shape embedding volume $V_{\rm{se}}$ with size $d_{\rm{se}}\times D_{\rm{se}} \times W_{\rm{se}} \times H_{\rm{se}}$, where $D_{\rm{se}}={D_{\rm{occ}}}/{b}$, $W_{\rm{se}}={W_{\rm{occ}}}/{b}$ and $H_{\rm{se}}={H_{\rm{occ}}}/{b}$.

\subsection{Differentiable Trilinear Sampling Module}

After generating the shape embedding volume $V_{\rm{se}}$, we need to obtain the pointwise shape embedding $\textbf{e}_i \in \mathbb{R}^{d_{\rm{se}}}$ for query point $p_i:\textbf{x}_i \in \Omega_0$. A straightforward way is to use the voxel-wise shape embedding $\textbf{s}_i \in \mathbb{R}^{d_{\rm{se}}}$ of the voxel center nearest to $p_i$. But to maintain the continuity of the latent shape field at the voxel borders, we use trilinear interpolation $f_{\rm{tri}}$ to sample $\textbf{e}_i$ for $p_i$ from its 8 nearest voxel centers.
\begin{equation}
f_{\rm{tri}}(\textbf{x}_i, V_{\rm{se}}) = \textbf{e}_i.
\end{equation}
Note that for points near the edge of $V_{\rm{se}}$ or when $H_{\rm{se}}$ equals $1$, the method degrades to bilinear sampling.

Formally, we first normalize the length of voxel edge. And then the trilinear sampling for $\textbf{e}_i$ can be written as:
\begin{equation}
\begin{split}
&e_i^c = \sum\limits_{m} ^{D_{\rm{se}}} \sum\limits_{n} ^{W_{\rm{se}}} \sum\limits_{k} ^{H_{\rm{se}}} s_{mnk}^c \times {\rm{max}}(0,1-|x_i-x_m|)\\&\times{\rm{max}}(0,1-|y_i-y_n|)\times{\rm{max}}(0,1-|z_i-z_k|).
\end{split}
\end{equation}
where $e_i^c$ and $s_{mnk}^c$ are the shape embedding values on channel $c$ for $p_i: \textbf{x}_i=(x_i, y_i, z_i) $ and voxel center $q_{mnk}:\textbf{x}_{mnk}=(x_m, y_n, z_k)$. Then we can obtain the gradients with respect to $\textbf{s}_{mnk}$ for backpropagation:
\begin{equation}
\begin{split}
&\frac{\partial e_i^c}{\partial  s_{mnk}^c} = \sum\limits_{m} ^{D_{\rm{se}}} \sum\limits_{n} ^{W_{\rm{se}}} \sum\limits_{k} ^{H_{\rm{se}}} {\rm{max}}(0,1-|x_i-x_m|)\\&\times{\rm{max}}(0,1-|y_i-y_n|)\times{\rm{max}}(0,1-|z_i-z_k|).
\end{split}
\end{equation}

This differentiable trilinear sampling mechanism allows loss gradients to flow back to the shape embeddings and further back to $f_{\rm{dis}}$, making it possible to train discriminative model and the following generative model cooperatively.


\subsection{Positional Encoding Module}
Positional encoding has proved a very effective technique in neural rendering\cite{c14} \cite{c34} for its capacity to capture high-frequency information. We leverage it in our formulation to  represent more geometric details of the signed distance field. Thus we encode the 3D Cartesian coordinates $\textbf{x}_i$ into high-dimensional features $\textbf{y}_i \in \mathbb{R}^{d_{\rm{enc}}}$. Specifically, the positional encoding function $f_{\rm{enc}}$ has the form like:
\begin{equation}
\textbf{y}_i := f_{\rm{enc}}(\textbf{x}_i) = (\gamma_{\rm{enc}}(x_i), \gamma_{\rm{enc}}(y_i), \gamma_{\rm{enc}}(z_i)).
\end{equation}
Applied to each component of $\textbf{x}_i$, the function $\gamma_{\rm{enc}}(\cdot)$ is a mapping from $\mathbb{R}$ to $\mathbb{R}^{2L}$:
\begin{equation}
\begin{split}
\gamma_{\rm{enc}}(p) = ({\rm{sin}}(2^0 \pi p), &{\rm{cos}}(2^0 \pi p), \cdots, \\
 &{\rm{sin}}(2^{L-1} \pi p), {\rm{cos}}(2^{L-1} \pi p)).
 \end{split}
\end{equation}
where L denotes the number of frequency octaves, and therefore $d_{\rm{enc}}=3\times 2L$.

\begin{figure*}[t]
\centerline{\includegraphics[width=1\textwidth]{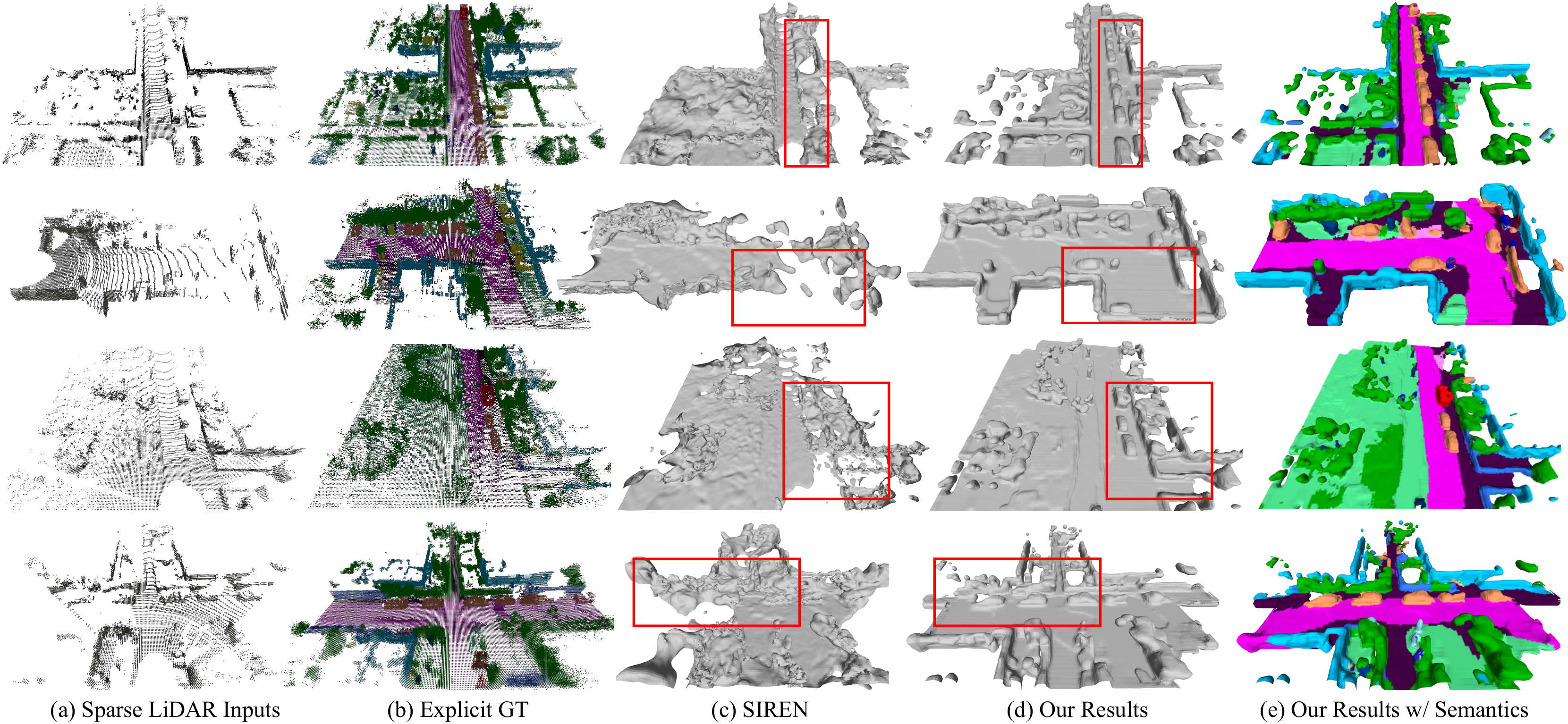}}
\caption{Qualitative results of implicit (semantic) road scene completion on the \emph{SemanticKITTI} validation set.}
\label{fig:qualitative}
\end{figure*}

\subsection{Generative Model}

We utilize the sinusoidal representation networks (SIREN) as the backbone of our generative model, which leverages sine as a periodic activation function for implicit neural representations. The function $\Phi(\textbf{x})$ represented by SIREN can be formalized as:
\begin{equation}
\begin{split}
\Phi(\textbf{x}) = &{\mathbf W}_n(\phi_{n-1} \circ \phi_{n-2} \circ \cdots \circ \phi_0)(\textbf{x})+{\mathbf b}_n, \\
&{\textbf{x}}_j \mapsto \phi_{j}({\textbf{x}}_j)={\rm{sin}}({\mathbf W}_j {\textbf{x}}_j + {\mathbf b}_j).
\end{split}
\end{equation}
where $\phi_{j}:\mathbb{R}^{M_j}\mapsto \mathbb{R}^{N_j}$ is the $j^{th}$ layer of SIREN. Given ${\textbf{x}}_j \in \mathbb{R}^{M_j}$, the layer applies the affine transform with weights ${\mathbf W}_j \in \mathbb{R}^{N_j \times M_j}$ and biases ${\mathbf b}_j \in \mathbb{R}^{N_j}$ on it, and then pass the resulting vector to the sine nonlinearity which operates on each component.

In our generative model $f_{\rm{gen}}$, we modify SIREN to be a conditional one. We take the concatenated encoded coordinates and pointwise shape embedding $[\textbf{y}_i, \textbf{e}_i]$ as input instead of the raw coordinates $\textbf{x}_i$, predicting SDF value:
\begin{equation}
f_{\rm{gen}}([\textbf{y}_i, \textbf{e}_i]) \approx {\rm{SDF}}(\textbf{x}_i).
\end{equation}
And the weights of our model are shared for all scenes. We leverage the proposed loss function \eqref{equ:sdfloss} to optimize model weights and shape embeddings. Note that during training, $\textbf{x}_i$ is sampled from dense ground truth $\Omega_1^\prime$ instead of sparse input $\Omega_1$. Therefore, in a data driven manner, our generative model can effectively map the shape embedding space to the signed distance space with abundant geometric information.


\subsection{Optional SSC part}

Furthermore, we extend our method in two ways for implicit semantic scene completion (SSC), as shown in the SSC Option A/B parts in Fig.\ref{fig:mainfig}.

\textbf{SSC Option A.} We add a dense discriminative head to predict SSC results which can be mapped to the implicit function using K-Nearest-Neighbor. Specifically, we first leverage a sparse Seg-Net similar to Com-Net for semantic segmentation of $V_{\rm{occ}}$. Then we use a dense convolutional network named SSC-Net to predict coarse semantic completion results. Mapping it to our representation, we get the refined implicit semantic results.

\textbf{SSC Option B.} We add a parallel implicit generative head to directly model the implicit semantic label field. Its structure is similar to our SDF generative model, except that it outputs the probabilities of label classification. 

We supervise the semantic segmentation and completion results with a multi-classification cross entropy loss:
\begin{equation}
\mathcal{L}_{\rm{seg}} = -{\frac 1 {N_{\rm{seg}}}}\sum_{i=1} ^ {N_{\rm{seg}}}\sum_{c=1} ^ {C} y_{i,c} {\rm{log}}(p_{i,c}).
\end{equation}
where $y_{i,c}$ and $p_{i,c}$ are the actual and predicted probability for point $i$ belonging to category $c$ respectively. $N_{\rm{seg}}$ points and $C$ categories are considered.

\subsection{Training and Inference}

During training, we randomly sample $N_{\rm{on}}$ points from $\Omega_1^\prime$ and $N_{\rm{off}}$ points from $\Omega_3^\prime$, optimizing the whole neural network with loss:
\begin{equation}
\begin{split}
\mathcal{L}_{\rm{total}} = \mathcal{L}_{\rm{SDF}} + \lambda_5\mathcal{L}_{\rm{com}} + \lambda_6\mathcal{L}_{\rm{seg}}.
\end{split}
\end{equation}
Here, $\lambda_5$ and $\lambda_6$ are constant weight parameters. Note that $\lambda_6 = 0$ when SSC part is not used.

During inference, we uniformly sample $N_{\rm{inf}}^3$ points from $\Omega_0$ at a specified resolution. And we use a threshold $v_{\rm {th}}$ close to zero to select explicit surface points from estimated SDF values for evaluation:
\begin{equation}
f_{\rm{sel}}(\textbf{x}) = \left\{
\begin{aligned}
1  & ,\ {\rm{if}} \ |f_{\rm{sdf}}(\textbf{x}, \Omega_1)| \leq v_{\rm {th}};\\
0  & ,\ {\rm{else}}.
\end{aligned}
\right.
\end{equation}
\begin{equation}
f_{\rm{sdf}}(\textbf{x}, \Omega_1) = f_{\rm{gen}}([f_{\rm{enc}}(\textbf{x}),f_{\rm{tri}}(\textbf{x}_i, f_{\rm{dis}}(f_{\rm{vox}}(\Omega_1)))]).
\end{equation}



%% file: tex/experiments.tex
\textbf{Dataset.} We evaluate our method on the \emph{SemanticKITTI} dataset \cite{c2}. There are a total of $22$ sequences ($8550$ scans) collected from the KITTI odometry dataset, in which $10$ sequences are used for training ($3834$ scans), $1$ sequence for validation ($815$ scans) and $11$ sequences for testing ($3901$ scans).  Each scan covers a range of $51.2$ m ahead of the car, $25.6$ m to each side and $6.4$ m in height. Every point in the completion ground truth has a semantic label out of $20$ classes, containing one \emph{unlabelled} class.

\textbf{Implementation Details.} For voxelization, we set $D_{\rm{occ}}=256$, $W_{\rm{occ}}=256$ and $H_{\rm{occ}}=32$. For Com-Net ($m = 5$), the convolutional kernel size and stride are $[2,2]$ for operations between two spatial scales, and $[3,1]$ otherwise. We set $\lambda_1=3000$, $\lambda_2=100$, $\lambda_3=100$, $\lambda_4=50$, $\lambda_5=100$ and use the Adam optimizer with an initial learning rate of $10^{-4}$. When the SSC module is included, we set $\lambda_6=50$. For the generative model, we use $N_{\rm{on}}=N_{\rm{off}}=16000$ and $N_{\rm{inf}}=256$. The ground truth point cloud in $\Omega_1^\prime$ for $f_{\rm gen}$ training is generated by accesssing the voxel centers of ground truth volumes and scaling them into the range of $[-1,1]^3$.


\subsection{Scene Completion}

In Table.\ref{tab:sc}, we compare our method with SIREN on \emph{SemanticKITTI} validation set. Directly comparing the input sparse point cloud with completion ground truth yields 10.268\% IoU. SIREN fitting improves the IoU to 26.256\%, which is still quite low despite a large relative margin. Thanks to the new dense boundary value formulation, our approach further improves IoU to 51.02\%.

\begin{table}[thpb]
\centering
\begin{tabular}{cccc}
\toprule
&Input  & SIREN  & Ours            \\
\midrule
IoU (\%) & 10.268 & 26.256 & \textbf{51.020} \\
\bottomrule
\end{tabular}
\vspace{1mm}
\caption{Scene completion results on the \emph{SemanticKITTI} validation set. Our approach outperforms SIREN by a large margin.}
\label{tab:sc}
\end{table}

This large improvement is better demonstrated with qualitative results in Fig.~\ref{fig:qualitative}.
Although SIREN is successful for clean uniform point cloud data, fitting large-scale outdoor scenes captured by LiDAR is much more difficult (Fig.\ref{fig:qualitative}-a). On the one hand, many points are not scanned thus missing due to occlusion. On the other hand, caused by the mechanism of LiDAR, data sparsity increases with distance and it is extremely sparse at the far end. This results in the lack of effective boundary values. For this reason, as a pure generative model, SIREN fails to fit road scenes and produces lots of artifacts (Fig.\ref{fig:qualitative}-c). Our method, on the contrary, leverages data-driven shape codes generated by a strong sparse convolutional network and successfully completes the scenes. As shown in Fig.\ref{fig:qualitative}-d and highlighted in red boxes, both occluded and incomplete shapes are better reconstructed than SIREN.

\begin{figure}[thpb]
\centerline{\includegraphics[width=0.35\textwidth]{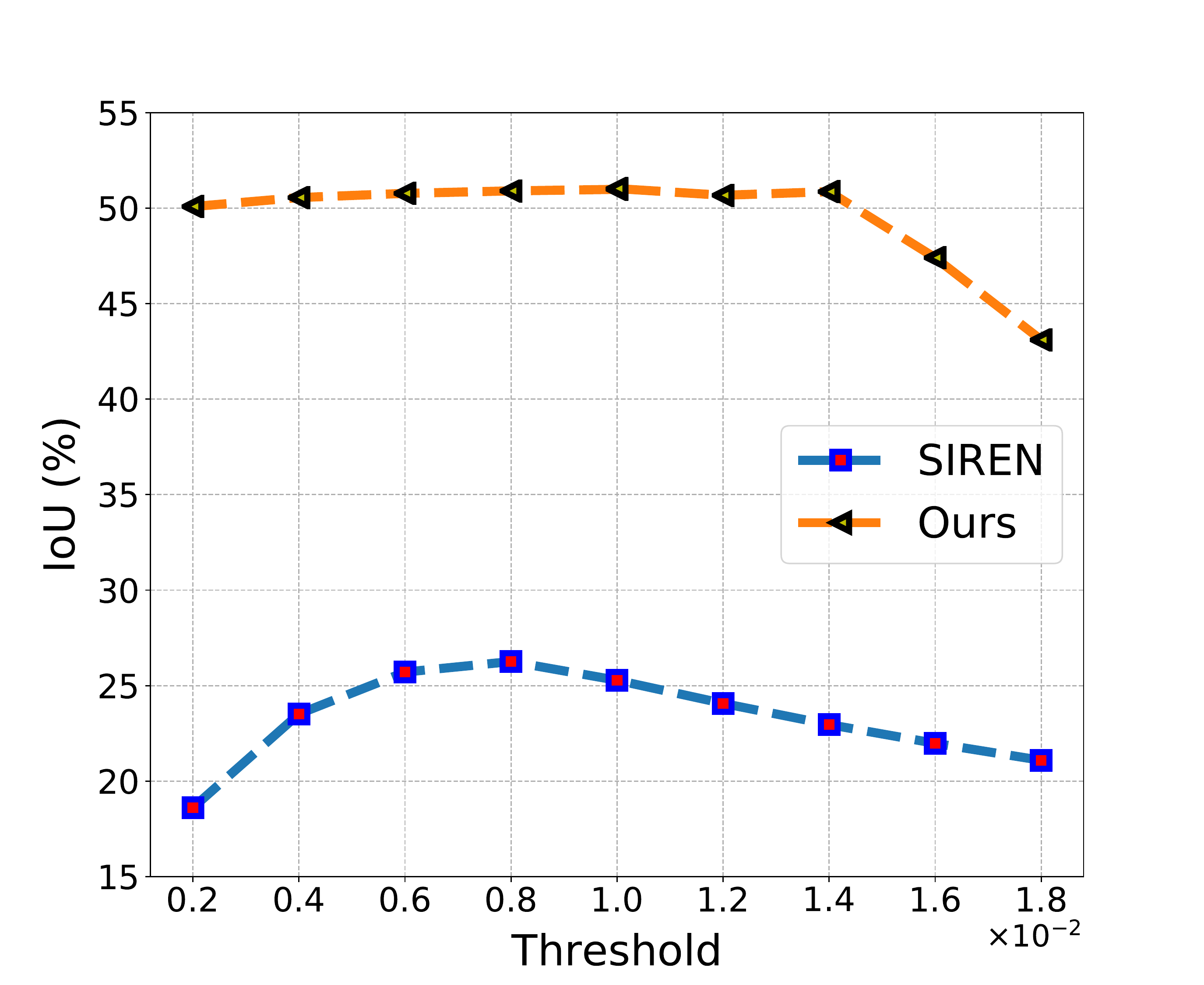}}
\caption{IoU comparisons under different thresholds.}
\label{fig:IoUcd}
\end{figure}


In order to show that the significant margins reported in Table.~\ref{tab:sc} are robust to Marching Cubes thresholds, we provide an exhaustive evaluation in Fig.\ref{fig:IoUcd}. It is clear that our method our-performs SIREN under all inspected thresholds.

\subsection{Ablation Study}

To better understand the newly proposed learning system, we provide a series of ablation studies as follows.

\textbf{The impact of data augmentation.} We randomly rotate scans along the gravity direction between $-45$ and $+45$ degrees. As shown in Table.\ref{tab:AblationDataAug}, this simple strategy improves IoU by 3.06\%. We also tried to further randomly flip scans by the symmetry plane of the car, whose impact is limited. With random rotation, the distribution of normal orientations becomes more uniform, which can help the shape embedding to better capture first-order boundary values.
\begin{table}[thpb]
\centering
\begin{tabular}{cc}
\toprule
Data Augmentation & IoU (\%)\\
\midrule
/                 & 47.963\\
Rotate            & \textbf{51.020}\\
Rotate \& Flip    & 50.856\\
\bottomrule
\end{tabular}
\vspace{1mm}
\caption{Ablation study for data augmentation.}
\label{tab:AblationDataAug}
\end{table}

\textbf{Discriminative model design.}  For the discriminative model, we investigate three factors: (1) Where to add pruning blocks; (2) Conv layer number in the output block that generates shape embeddings; (3) Activation functions. Results are summarized in Table.\ref{tab:dis_model}, which demonstrate that our method is robust to these design choices.  
\begin{table}[thpb]
\centering
\begin{tabular}{cccc}
\toprule
\begin{tabular}[c]{@{}c@{}}Pruning\\ Blocks\end{tabular} & \begin{tabular}[c]{@{}c@{}}Output\\ Block\end{tabular} & Activations & IoU (\%)  \\
\midrule
Last 1                                                   & 2 convs                                                 & ELU         & 49.479          \\
Last 2                                                   & 2 convs                                                 & ELU         & 49.073          \\
Last 3                                                   & 2 convs                                                 & ELU         & 50.628          \\
Last 4                                                   & 2 convs                                                 & ELU         & 51.011          \\
All                                                      & 2 convs                                                 & ELU         & \textbf{51.020} \\
All                                                      & 4 convs                                                 & ELU         & 50.938          \\
All                                                      & 2 convs                                                 & ReLU        & 50.874          \\
\bottomrule
\end{tabular}
\vspace{1mm}
\caption{Ablation study for disriminative model.}
\label{tab:dis_model}
\end{table}

\textbf{Does generative model capacity matter?} Although deeper and wider models usually achieve better results for recognition, whether model capacity matters for our generative model remains an open question. We modify the width, depth and activation function of the multilayer perceptron. As shown in Table.\ref{tab:gen_model}, different configurations produce similar results. It demonstrates the capacity of generative model is not a performance bottleneck. Interestingly, using ReLU instead of Sine activation only brings a performance drop of 1.75\%. It suggests that in challenging scenarios like ours, using Sine acitvation is not as critical as in SIREN. 
\begin{table}[thpb]
\centering
\begin{tabular}{cccc}
\toprule
Width & Depth & Activations & IoU (\%) \\
\midrule
128   & 4     & Sine        & \textbf{51.038} \\
256   & 4     & Sine        & 51.020   \\
512   & 4     & Sine        & 50.932   \\
256   & 3     & Sine        & 49.603   \\
256   & 5     & Sine        & 50.910   \\
256   & 4     & ReLU        & 49.267   \\
\bottomrule
\end{tabular}
\vspace{1mm}
\caption{Ablation study for generative model capacity.}
\label{tab:gen_model}
\end{table}

\textbf{Which dimension of shape volume matter?}
To study which factor is the deciding one for the representation power of the shape volume, we evaluate different shape embedding dimensions and scale sizes. By scale size, we mean the cube size $b$ of the Com-Net. The results are shown in Table.\ref{tab:DimensionScale}, showing that using shape embeddings of dimension 128 is already capable of representing our scenes well. But increasing scale size leads to a sharp drop of IoU.
\begin{table}[thpb]
\centering
\begin{tabular}{ccc}
\toprule
\begin{tabular}[c]{@{}c@{}}Shape\\ Dimension\end{tabular} & \begin{tabular}[c]{@{}c@{}}Scale\\ Size\end{tabular} & IoU (\%)\\
\midrule
128&4&50.910\\
512&4&\textbf{51.193}\\
256&2&50.267\\
256&4&51.020\\
256&8&49.186\\
256&16&44.818\\
256&32&39.716\\
\bottomrule
\end{tabular}
\vspace{1mm}
\caption{Ablation study for shape embedding.}
\label{tab:DimensionScale}
\end{table}

\textbf{Is trilinear sampling necessary?} We justify the necessity of trilinear sampling in our method using Table.~\ref{tab:trilinear}. A trivial nearest neighbor sampling leads to a performance drop of 2.88\%. This is a clear margin that shows the benefit of smoothly interpolating shape embeddings. 

\begin{table}[thpb]
\centering
\begin{tabular}{cc}
\toprule
Sample Strategy & IoU (\%) \\
\midrule
Trilinear&\textbf{51.020}\\
Nearest&48.144\\

\bottomrule
\end{tabular}
\vspace{1mm}
\caption{Ablation study for sampling strategy.}
\label{tab:trilinear}
\end{table}

\begin{table*}[t]
  \centering
  \ra{1.05}
\resizebox{1\linewidth}{!}
 {
\small
\begin{tabular}{@{}l@{\hspace{2.7mm}}*{20}{c@{\hspace{2.7mm}}}c@{\hspace{2.7mm}}r@{\hspace{0.2mm}}r@{}}
  \toprule
 Approach&\rotatebox{90}{IoU (\%)}&\rotatebox{90}{mIoU (\%)}&\rotatebox{90}{car}&\rotatebox{90}{bicycle}&\rotatebox{90}{motorcycle}&\rotatebox{90}{truck}&\rotatebox{90}{other-vehicle}&\rotatebox{90}{person}&\rotatebox{90}{bicyclist}&\rotatebox{90}{motorcyclist}&\rotatebox{90}{road}&\rotatebox{90}{parking}&\rotatebox{90}{sidewalk}&\rotatebox{90}{other-ground}&\rotatebox{90}{building}&\rotatebox{90}{fence}&\rotatebox{90}{vegetation}&\rotatebox{90}{trunk}&\rotatebox{90}{terrain}&\rotatebox{90}{pole}&\rotatebox{90}{traffic-sign}\\
 \hline

 Our SSC-A&50.1&20.2&39.1&1.11&4.99&25.2&17.0&4.56&2.43&0&64.5&21.6&36.5&\textbf{3.58}&29.1&12.4&35.5&18.0&42.2&17.2&9.47\\

 Our SSC-B&50.8&18.0&37.7&0.82&3.60&16.9&9.85&3.76&1.33&0&64.3&14.7&34.5&3.04&28.8&10.5&34.1&15.5&42.1&14.8&5.22\\

 JS3C-Net&\textbf{53.1}&22.7&40.5&10.5&12.1&28.2&15.8&\textbf{8.86}&\textbf{2.59}&0&58.4&23.3&37.2&1.66&\textbf{36.4}&13.8&\textbf{40.7}&21.5&\textbf{47.6}&20.3&11.4\\
 
 Ours w/ JS3C&51.0&\textbf{23.4}&\textbf{41.7}&\textbf{10.6}&\textbf{12.6}&\textbf{33.6}&\textbf{18.3}&8.12&2.36&0&\textbf{64.8}&\textbf{26.0}&\textbf{39.4}&2.07&30.4&\textbf{14.7}&37.6&\textbf{22.7}&46.0&\textbf{20.8}&\textbf{13.6}\\
 
 \bottomrule
  \end{tabular}}
  \vspace{1mm}
  \caption{Semantic scene completion results on the \emph{SemanticKITTI} validation set.} 
  \label{tab:segmentation}
\end{table*}

\begin{figure}[thpb]
\centerline{\includegraphics[width=0.45\textwidth]{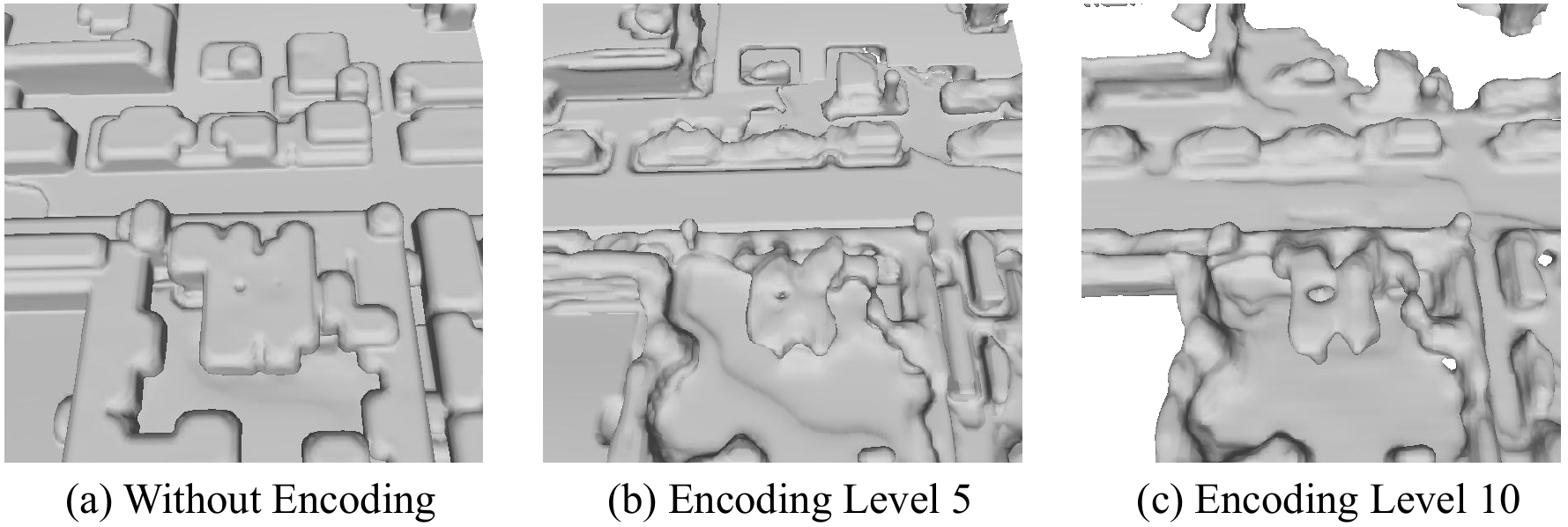}}
\caption{Results with different positional encoding strategies.}
\label{fig:encoding_level}
\end{figure}

\textbf{How to encode positional information?}
The goal of positional encoding module is to represent fine geometric details of the scene. We investigate positional encoding levels and whether to concatenate original coordinates. For example, setting positional encoding level L to 10 and concatenating the original 3D coordinates yields a 63-dimension representation. Results are summarized in Table.\ref{tab:AblationPosEncoding}. When positional encoding is not used or the encoding level is low, the completion IoU decreases dramatically. Through the qualitative results in Fig.~\ref{fig:encoding_level}, it is clear that leaving out postional encoding leads to the loss of details.
\begin{table}[t]
\centering
\begin{tabular}{cccc}
\toprule
\begin{tabular}[c]{@{}c@{}}Positional\\ Encoding\end{tabular} & \begin{tabular}[c]{@{}c@{}}Include\\ xyz\end{tabular} & \begin{tabular}[c]{@{}c@{}}Encoding\\ Level\end{tabular} & IoU (\%) \\
\midrule
{$\bm{\times}$}&-&-&40.431\\
\checkmark&\checkmark&5&40.331\\
\checkmark&\checkmark&10&51.020\\
\checkmark&\checkmark&15&50.871\\
\checkmark&{$\bm{\times}$}&10&\textbf{51.075}\\
\bottomrule
\end{tabular}
\vspace{1mm}
\caption{Ablation study for positional encoding.}
\label{tab:AblationPosEncoding}
\end{table}

\subsection{Semantic Scene Completion}
Table.\ref{tab:segmentation} shows semantic scene completion results on the \emph{SemanticKITTI} validation set, which is evaluated on 19 categories. The SSC-A and SSC-B designs achieve 20.2\% and 18.0\% mIoU, respectively. Although they under-perform the state-of-the-art method JS3C-Net \cite{c8}, our models allow implicit completion and are trained in a semi-supervised manner. Qualitative results (drawn with SSC-A) shown in Fig.~\ref{fig:qualitative} demonstrate faithful semantic implicit completion. Last but not least, we map explicit semantic completion results from JS3C-Net to our implicit completion results using K-Nearest-Neighbors, achieving 23.4\% mIoU.


\begin{figure}[thpb]
\centerline{\includegraphics[width=0.45\textwidth]{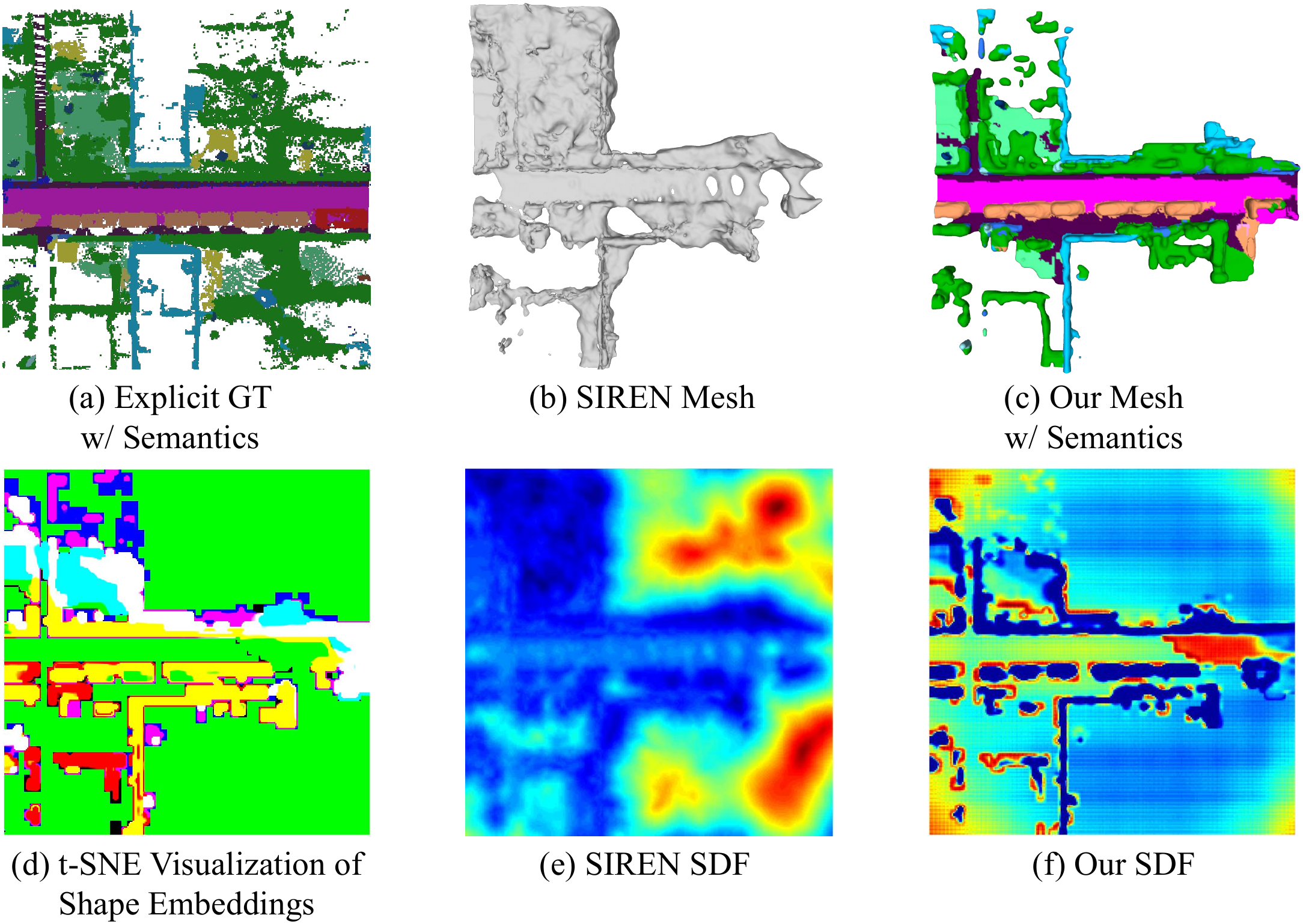}}
\caption{We analyze a slice parallel to the ground from scene (a). (b) and (c) are meshes reconstructed by SIREN and our method. (d) is the t-SNE visualization of shape embeddings generated by our model. (e) and (f) visualize the SDF field of the slice.}
\label{fig:data_vis}
\end{figure}

\subsection{Visualization}

\textbf{Shape embedding Field.} To probe the shape embedding space learned by our model, we leverage t-SNE to reduce its dimension to 3, and visualize them as RGB values. As shown in Fig.\ref{fig:data_vis}-d, clear clusters and sharp boundaries demonstrate that our shape embedding space well encodes both zeroth-order and first-order boundary values. 

\textbf{SDF Field.} As shown in Fig.\ref{fig:data_vis}-e/f, the SDF field generated by our method is much more consistent with the underlying scene than SIREN, although top-right and bottom-right regions are equally difficult for both methods. 

\begin{figure}[t]
\centerline{\includegraphics[width=0.5\textwidth]{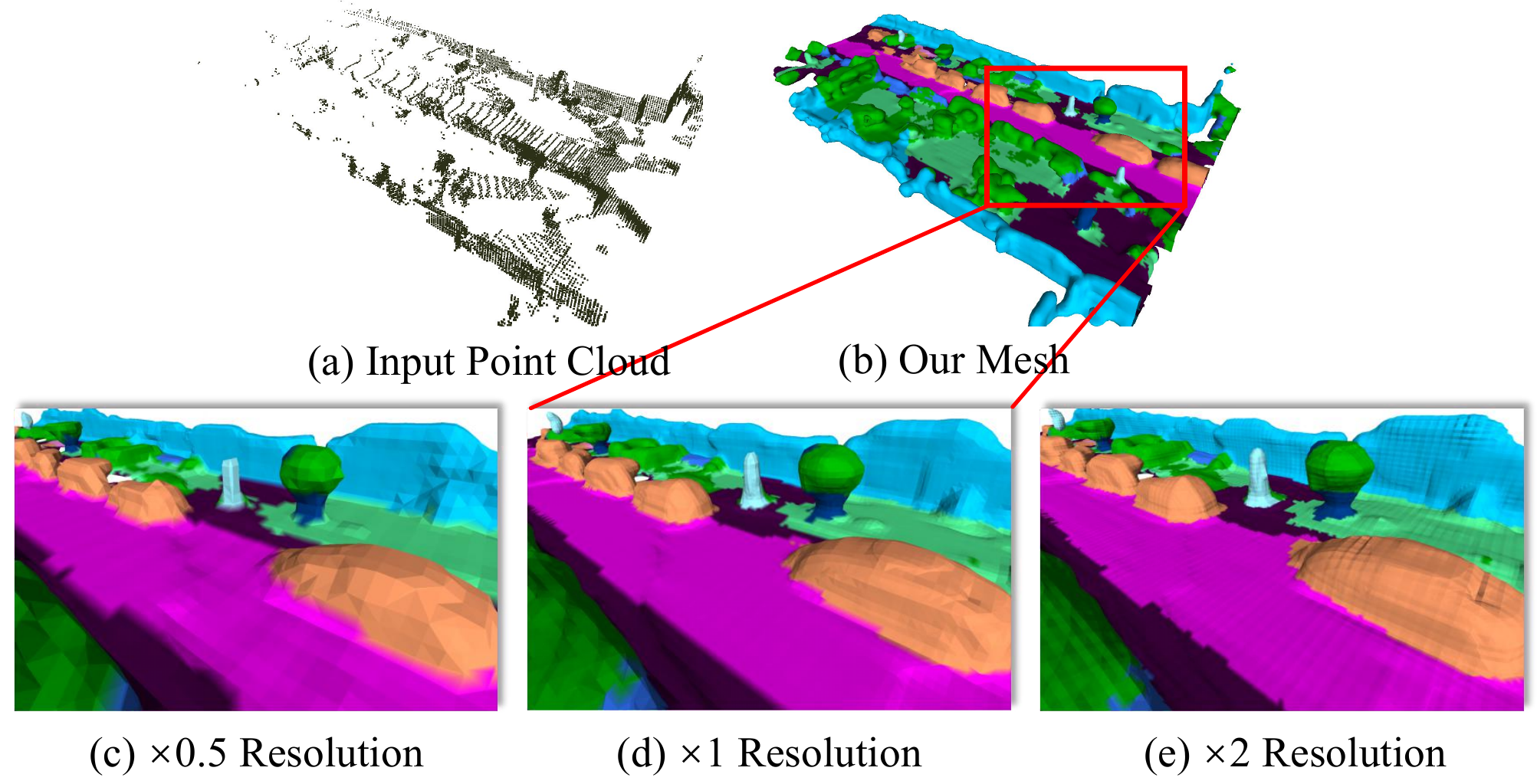}}
\caption{Scene completion results at multiple resolutions.}
\label{fig:multi_resolution}
\end{figure}

\textbf{Arbitrary resolution.}
Our formulation represents the whole scene with its continuous signed distance field. As such, we can get mesh reconstructions at any resolution. As shown in Fig.\ref{fig:multi_resolution}, we generate the mesh at three resolutions, where $\times 1$ means the same resolution as input point cloud. 

%% file: tex/conclusion.tex
In this work, we propose a novel semi-supervised formulation for implicit scene representation. Learned shape embeddings are treated as dense boundary values that constraint semi-supervised signed distance function learning. We implement the formulation as a hybrid neural network combining discriminant and generative models. The network is trained to implicitly fit road scenes captured by sparse LiDAR point clouds, without accessing exact SDF values at free space. Large-scale evaluations on the \emph{SemanticKITTI} dataset show that our method out-performs SIREN by a large margin. We also extend the proposed method for semantic implicit completion in two ways, achieving strong qualitative and quantitative results.

\textbf{Limitations.} Theoretically, this semi-supervised formulation allows test-time fune-tuning on target scans. However, we have not obtained positive results thus far.

%% file: tex/supplement.tex
\subsection{More Architecture Details}

\subsubsection{Discriminative Model}

\textbf{Sparse convolutional network.} Because the input occupancy volume $V_{\rm{occ}}$ is extremely sparse, to reduce memory requirement, we employ the Minkowski Engine \cite{c12} to build our Com-Net. Unlike dense convolutional operation imposed on data volume, the sparse counterpart in Minkowski Engine only needs to access the coordinates and features of non-empty voxels stored in a hash table, resulting in a sparse tensor $\mathbf{x} = [\mathbf C_{n \times d}, \mathbf F_{n \times m}]$, where $n$ denotes the number of non-empty voxels, $d$ and $m$ denote the dimension of coordinates and features. Then, the generalized sparse convolution in D-dimension has the form:
\begin{equation}
\mathbf{x}_\mathbf{u}^\mathtt{out} = \sum\limits_{\mathbf i \in \mathcal{N}^D(\mathbf{u}, K, \mathcal{C}^\mathtt{in})} \mathbf{W}_\mathbf{i} \mathbf{x}_\mathbf{u+i}^\mathtt{in} \ \rm{for} \ \mathbf{u} \in \mathcal{C}^\mathtt{out}.
\end{equation}
Where $K$ is convolution kernel size and $\mathcal{N}^D$ is a set of offsets which are at most $\lceil \frac{1}{2}(K-1) \rceil$ voxels away from a voxel center of interest $\mathbf{u}$. $\mathbf{W}_\mathbf{i}$ is the weight matrix for offset $\mathbf{i}$. $\mathcal{C}^\mathtt{in}$ and $\mathcal{C}^\mathtt{out}$ are input and output coordinates of sparse tensors.

As such, we extract the coordinates and features from $V_{\rm{occ}}$, generating $\mathbf{x}_{\rm{occ}} = [\mathbf C_{N_{\rm{occ}} \times 3}, \mathbf F_{N_{\rm{occ}} \times 1}]$ as Com-Net input. The output of Com-Net is a shape embedding volume $V_{\rm{se}}$ in the form $\mathbf{x}_{\rm{se}} = [\mathbf C_{N_{\rm{se}} \times 3}, \mathbf F_{N_{\rm{se}} \times d_{\rm{se}}}]$. $N_{\rm{occ}}$ and $N_{\rm{se}}$ denote the count of non-empty voxels in $V_{\rm{occ}}$ and $V_{\rm{se}}$ respectively. With the sparse-to-dense conversion in Minkowski Engine, we can get $V_{\rm{se}}$ in the dense form.

\textbf{Implementation Details.} Our Com-Net predicts shape embeddings via a shape completion process. It has a six-tiered design, which learns local geometry representation of scene information at different scales: $[1, \frac{1}{2}, \frac{1}{4}, \frac{1}{8}, \frac{1}{16}, \frac{1}{32}]$ of the original input volume size. The channel count of features at each tier is $[16,32,64,128,256,512]$ respectively. In the encoder part, each tier includes a convolutional block to increase feature dimension and two residual blocks to enhance representation learning. Each of them consists of two convolutional layers. In the decoder part, each tier includes a generative deconvolutional block capable of generating new voxels and two residual blocks. To get better shape completion results and avoid the \emph{submanifold dilation problem} \cite{c11}, we use a pruning block to prune off redundant voxels at each decoder tier. It includes a convolutional layer outputting the binary classification results of whether a voxel should be pruned, and a pruning operation based on the results: the voxels with positive output are kept and otherwise pruned. At the end of the decoder, a convolutional block is leveraged to raise the feature dimension to the preset shape embedding dimension $d_{\rm{se}}$. Then we use an average pooling layer to aggregate the features of voxels within a cube of size $b^3$ into a single one to get the final shape embedding output $\mathbf{x}_{\rm{se}}$. In addition, between each corresponding parts of encoder and decoder, we use skip-connections to utilize low-level information.

\subsubsection{Differentiable Trilinear Sampling Module}

\begin{figure}[tpb]
\centerline{\includegraphics[width=0.45\textwidth]{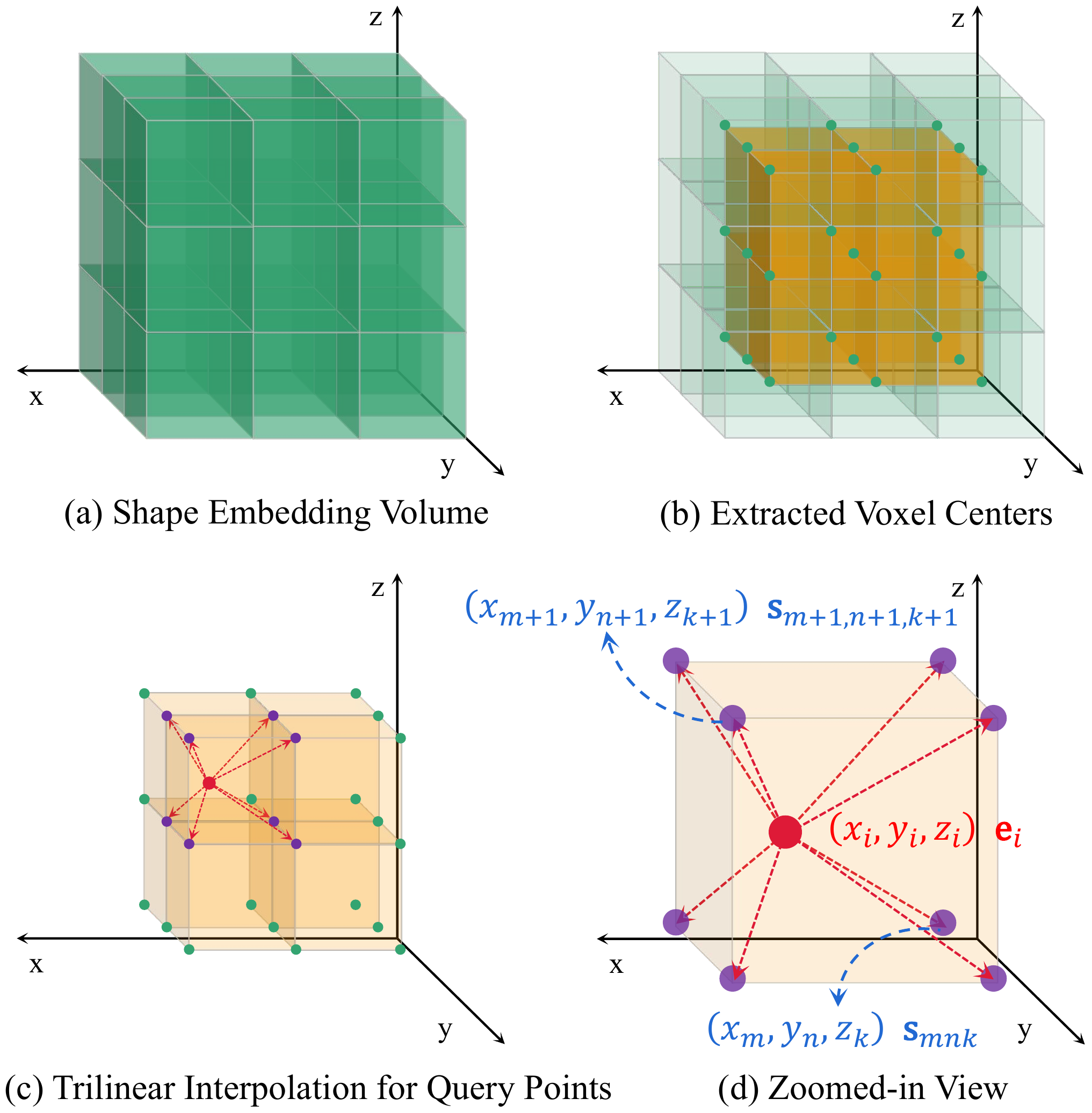}}
\caption{Differentiable Trilinear Sampling.}
\label{fig:trilinear}
\end{figure}

To maintain the continuity of the latent shape field, we use trilinear interpolation to sample the pointwise shape embedding $\textbf{e}_i \in \mathbb{R}^{d_{\rm{se}}}$. The process is illustrated in Fig.~\ref{fig:trilinear}. We first extract voxel centers from the shape embedding volume $V_{\rm{se}}$. Then we select the 8 voxel centers nearest to the query point $p_i$ and get the interpolation result $\textbf{e}_i$ from their voxel-wise shape embedding $\textbf{s}_i \in \mathbb{R}^{d_{\rm{se}}}$.

This trilinear sampling process for pointwise shape embedding $\textbf{e}_i$ can be formulated as:
\begin{equation}
\begin{split}
&e_i^c = \sum\limits_{m} ^{D_{\rm{se}}} \sum\limits_{n} ^{W_{\rm{se}}} \sum\limits_{k} ^{H_{\rm{se}}} s_{mnk}^c \times {\rm{max}}(0,1-|x_i-x_m|)\\&\times{\rm{max}}(0,1-|y_i-y_n|)\times{\rm{max}}(0,1-|z_i-z_k|).
\end{split}
\end{equation}
where $e_i^c$ and $s_{mnk}^c$ are the shape embedding values on channel $c$ for $p_i: \textbf{x}_i=(x_i, y_i, z_i) $ and voxel center $q_{mnk}:\textbf{x}_{mnk}=(x_m, y_n, z_k)$.

\subsubsection{Generative Model}

\begin{figure}[tpb]
\centerline{\includegraphics[width=0.45\textwidth]{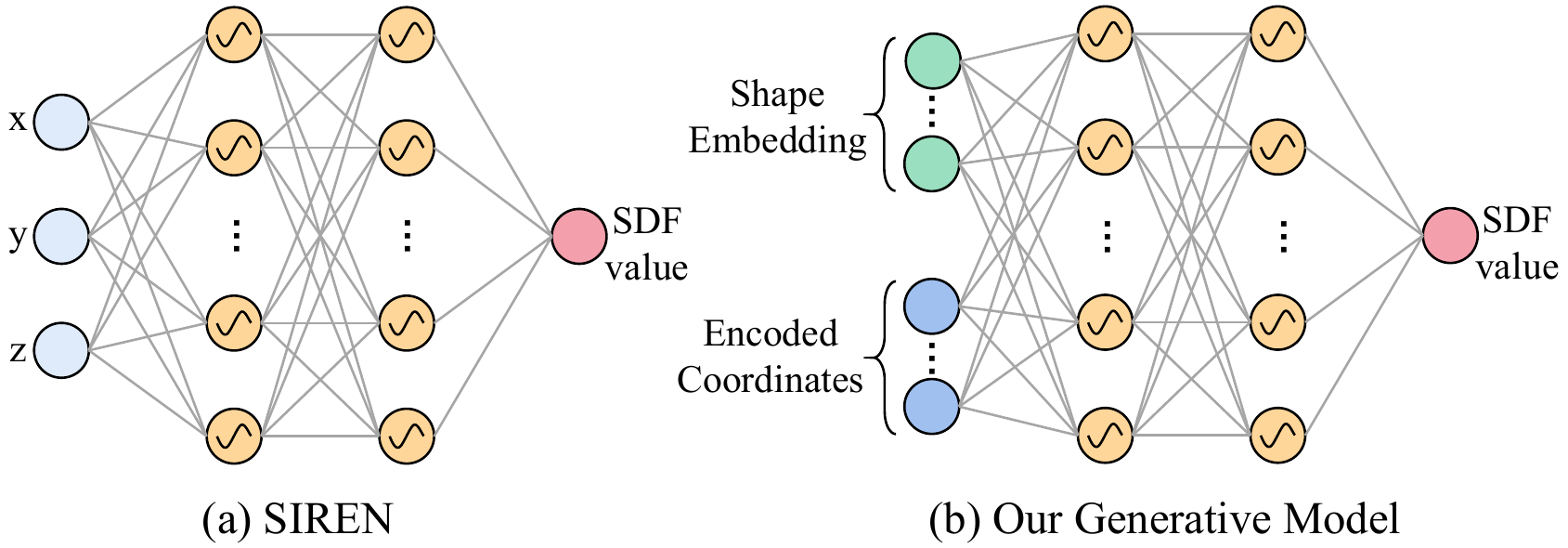}}
\caption{The generative network of SIREN and ours.}
\label{fig:generative_model}
\end{figure}

The original SIREN \cite{c10} takes the 3D-Cartesian coordinates as input. Instead, our generative model uses the concatenation of encoded coordinates and pointwise shape embedding as input. We use the same sinusoidal activation function and initialization scheme as SIREN. These two network architectures are shown in Fig.~\ref{fig:generative_model}.

\subsubsection{Optional SSC part}

\textbf{SSC Option A.}
We propose to use a sparse Seg-Net and a dense SSC-Net as an option for semantic scene completion. As such, the Seg-Net $f_{\rm{seg}}$ first maps the input $V_{\rm{occ}}$ to the categorical probability volume $V_{\rm{cat}}$ with size $C \times D_{\rm{occ}} \times W_{\rm{occ}} \times H_{\rm{occ}}$ in a sparse form $\mathbf{x}_{\rm{cat}} = [\mathbf C_{N_{\rm{occ}} \times 3}, \mathbf F_{N_{\rm{occ}} \times C}]$, where $C$ is the count of categories. Then, to leverage the semantic information for better completion, the Com-Net $f_{\rm{com}}$ is modified to take $V_{\rm{cat}}$ as input instead of $V_{\rm{occ}}$.
\begin{equation}
f_{\rm{seg}}(V_{\rm{occ}}) = V_{\rm{cat}}.
\end{equation}
\begin{equation}
f_{\rm{com}}(V_{\rm{cat}}) = V_{\rm{se}}.
\end{equation}
Meanwhile, the dense SSC-Net $f_{\rm{ssc}}$ takes $V_{\rm{cat}}$ as input and outputs coarse SSC result $V_{\rm{ssc}}$ with size $C \times D_{\rm{ssc}} \times W_{\rm{ssc}} \times H_{\rm{ssc}}$, where $D_{\rm{ssc}}=D_{\rm{occ}}$, $W_{\rm{ssc}}=W_{\rm{occ}}$ and $H_{\rm{ssc}}=H_{\rm{occ}}$.
\begin{equation}
f_{\rm{ssc}}(V_{\rm{cat}}) = V_{\rm{ssc}}.
\end{equation}
Mapping $V_{\rm{ssc}}$ to our signed distance representation, we can get the refined implicit semantic results. Specifically, we use K-Nearest-Neighbor to search the nearest labeled voxels in $V_{\rm{ssc}}$ for surface points extracted from estimated signed distance field. Then we assign the labels of these voxels to the corresponding surface points.

Built with Minkowski Engine, the architecture of Seg-Net is similar to Com-Net. Seg-Net has 5 tiers with channels [32,64,128,256,512]. Because there is no need to generate new non-empty voxels, we replace the generative deconvolution in the decoder part with the normal deconvolution, and remove the pruning blocks. We use a convolutional block at the end of the decoder to output classification result, which is supervised by a multi-classification cross entropy loss:
\begin{equation}
\mathcal{L}_{\rm{seg}} = -{\frac 1 {N_{\rm{occ}}}}\sum_{i=1} ^ {N_{\rm{occ}}}\sum_{c=1} ^ {C} y_{i,c} {\rm{log}}(p_{i,c}).
\end{equation}
where $y_{i,j}$ and $p_{i,j}$ are the true and predicted probability for voxel $i$ belonging to category $j$ respectively.

Then we modify the SSC module proposed in JS3C-Net \cite{c8} as our SSC-Net. It first uses a convolutional block to reduce the resolution, and then four convolutional blocks with residual connections are employed. After that, other four convolutional blocks with different scales are leveraged. Their outputs are concatenated together and fed into a convolutional block to get SSC result. Finally, the dense upsampling is used to restore the original resolution and output final result $V_{\rm{ssc}}$.

\textbf{SSC Option B.}
Alternatively, we use a parallel implicit generative head to directly model the semantic label field. It takes the same features as our SDF generative model as input and outputs the label classification probabilities.
\begin{equation}
f_{\rm{ssc}}([\textbf{y}_i, \textbf{e}_i]) \approx {\rm{Label}}(\textbf{x}_i).
\end{equation}
The result is supervised by loss:
\begin{equation}
\mathcal{L}_{\rm{seg}} = -{\frac 1 {{N_{\rm{on}}}+ {N_{\rm{off}}}}}\sum_{i=1} ^ {{N_{\rm{on}}}+ {N_{\rm{off}}}}\sum_{c=1} ^ {C} y_{i,c} {\rm{log}}(p_{i,c}).
\end{equation}
where $N_{\rm{on}}$ and $N_{\rm{off}}$ denote the counts of sampled points for generative models from $\Omega_1^\prime$ and $\Omega_3^\prime$ respectively.

\subsection{More Experiments}
\textbf{Input representation for the discriminative model.}
Because the sparsity of LiDAR data increases with radial distance and at different heights the sparsity is similar but the shape can wildly differ from each other, the radial distance and height may influence the result of our discriminative model. Therefore, we modify the input representation for our discriminative model. As shown in Table.\ref{tab:AblationFeatures}, these enhanced representations do provide better results, but the margins are small.

\begin{table}[thpb]
\centering
\begin{tabular}{cc}
\toprule
Input Features & IoU (\%)\\
\midrule
Occupancy                   & 51.020\\
Radial Distance             & 51.076\\
Radial Distance \& Height        & \textbf{51.120}\\
\bottomrule
\end{tabular}
\vspace{1mm}
\caption{Ablation study for input representation.}
\label{tab:AblationFeatures}
\end{table}

\textbf{Point sampling strategy for generative model.}
During training, we randomly sample $N_{\rm{on}}$ points from $\Omega_1^\prime$ and $N_{\rm{off}}$ points from $\Omega_3^\prime$ for the generative part. Specifically, the $N_{\rm{on}}$ on-surface points are randomly sampled from the occupied voxel centers of ground truth volumes. As for the $N_{\rm{off}}$ off-surface points, $N_{\rm{off}}/2$ of them are randomly sampled from the empty voxel centers of ground truth volumes, and the other $N_{\rm{off}}/2$ points are randomly sampled from the whole free space according to a uniform distribution.

As shown in Table.\ref{tab:AblationSample}, we evaluate different sampling strategies. For the experiment in the third row, in the set of off-surface points, we additionally use a group of points near the surface. These points are generated by adding small offsets to the on-surface points along the estimated normal directions (referred to as \emph{prior}). The results show that using more on-surface points helps training. And the manually generated nearby points bring a small drop of IoU, showing that they may give wrong zero-order boundary values.
\begin{figure*}[b]
\centerline{\includegraphics[width=1\textwidth]{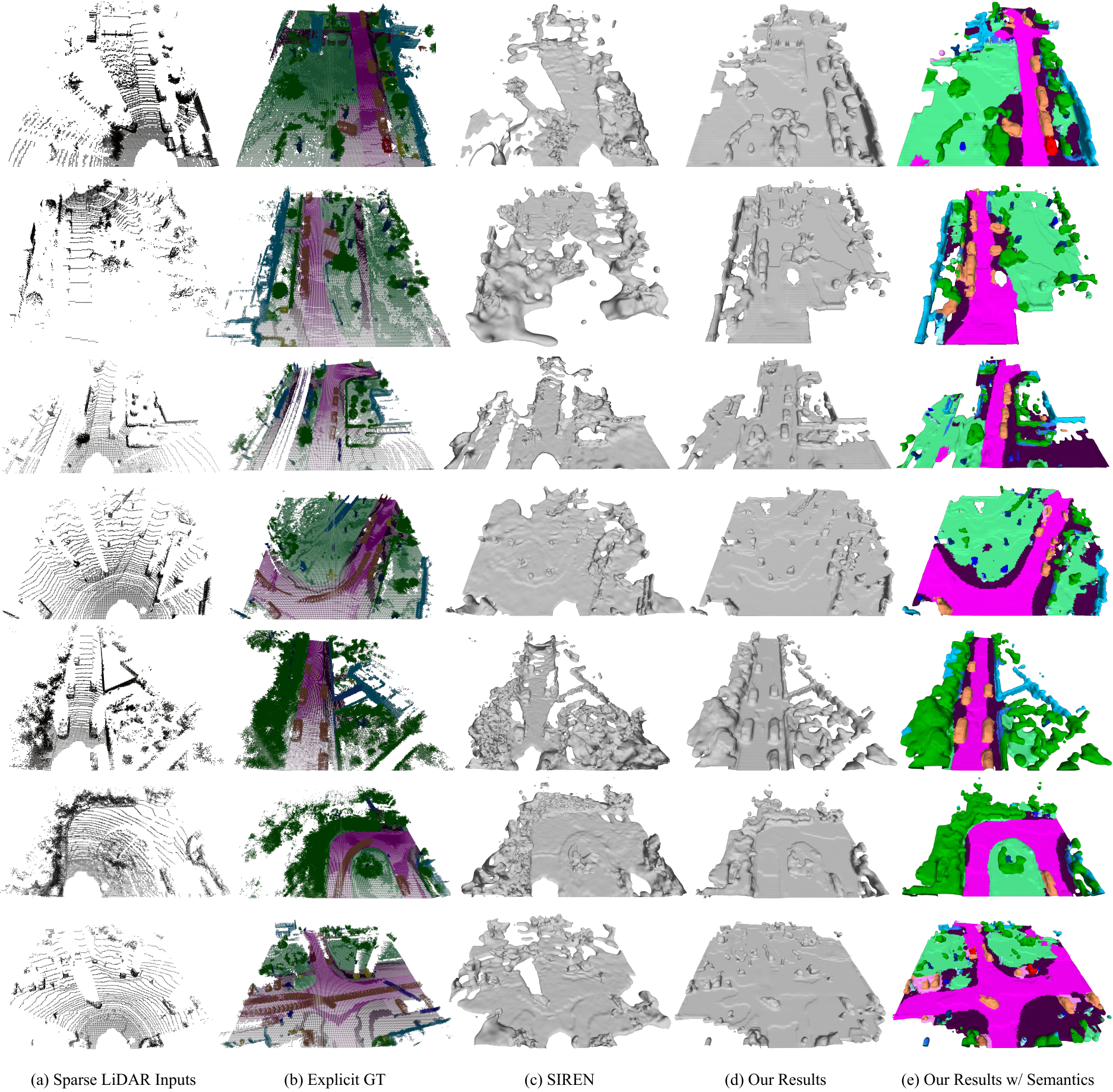}}
\caption{More qualitative results.}
\label{fig:more_qualitative_1}
\end{figure*}

\begin{table}[thpb]
\centering
\begin{tabular}{ccc}
\toprule
Points Count & Sampling Strategy & IoU (\%)\\
\midrule
$N_{\rm{on}}=N_{\rm{off}}$       &  random   & \textbf{51.020}\\
$N_{\rm{on}}=1/2 N_{\rm{off}}$  & random      & 50.395\\
$N_{\rm{on}}=2/3 N_{\rm{off}}$ & prior & 48.587\\
\bottomrule
\end{tabular}
\vspace{1mm}
\caption{Ablation study for point sampling strategy.}
\label{tab:AblationSample}
\end{table}

\subsection{More Qualitative Results}
We show more qualitative results in Fig.~\ref{fig:more_qualitative_1}. We select diverse road scenes for visualization. These results demonstrate that our method is robust to complex scenes.

\balance

\begin{figure*}[htpb]
\ContinuedFloat
\centerline{\includegraphics[width=1\textwidth]{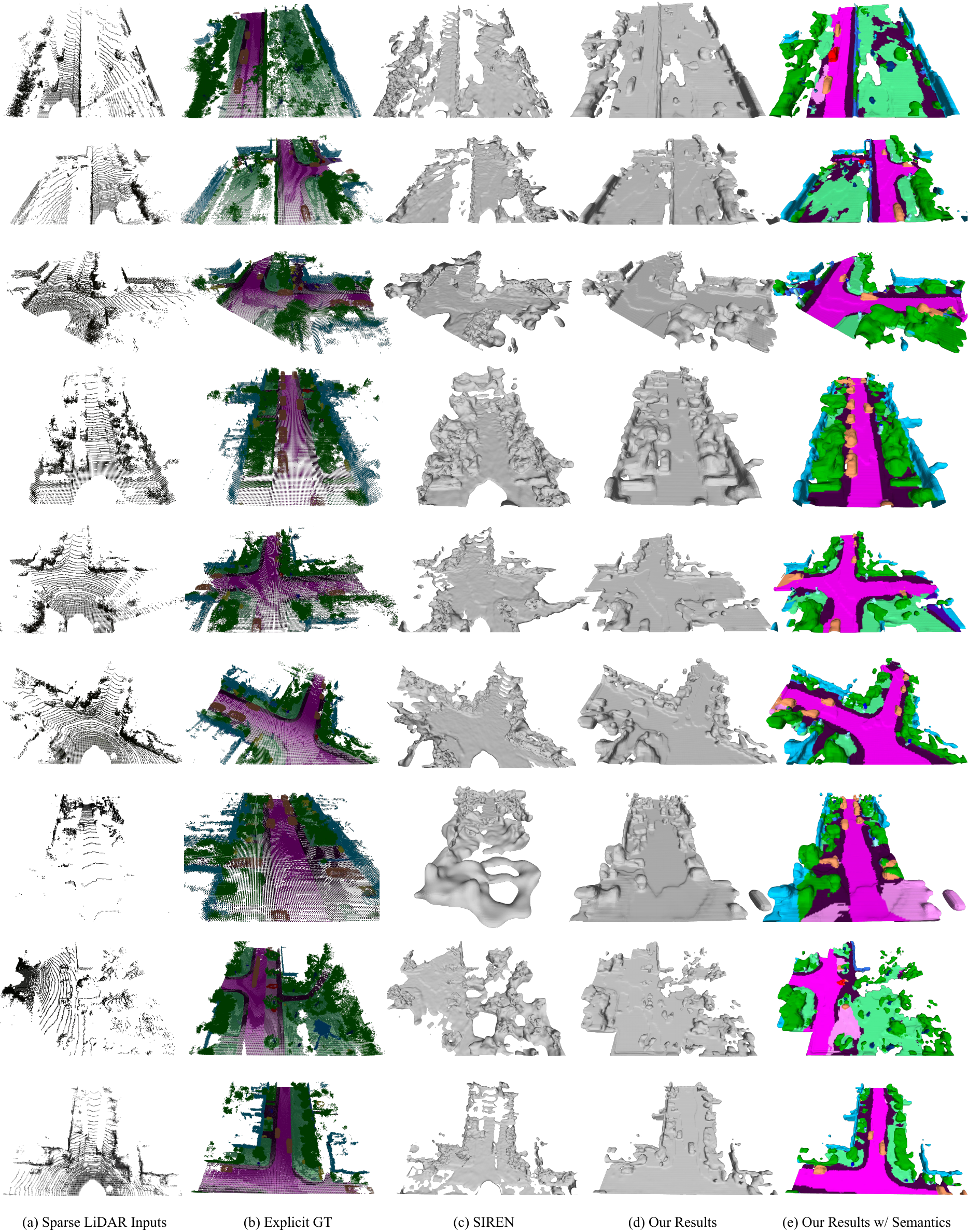}}
\caption{More qualitative results (cont.).}
\label{fig:more_qualitative_1}
\end{figure*}

\begin{figure*}[t]
\ContinuedFloat
\centerline{\includegraphics[width=1\textwidth]{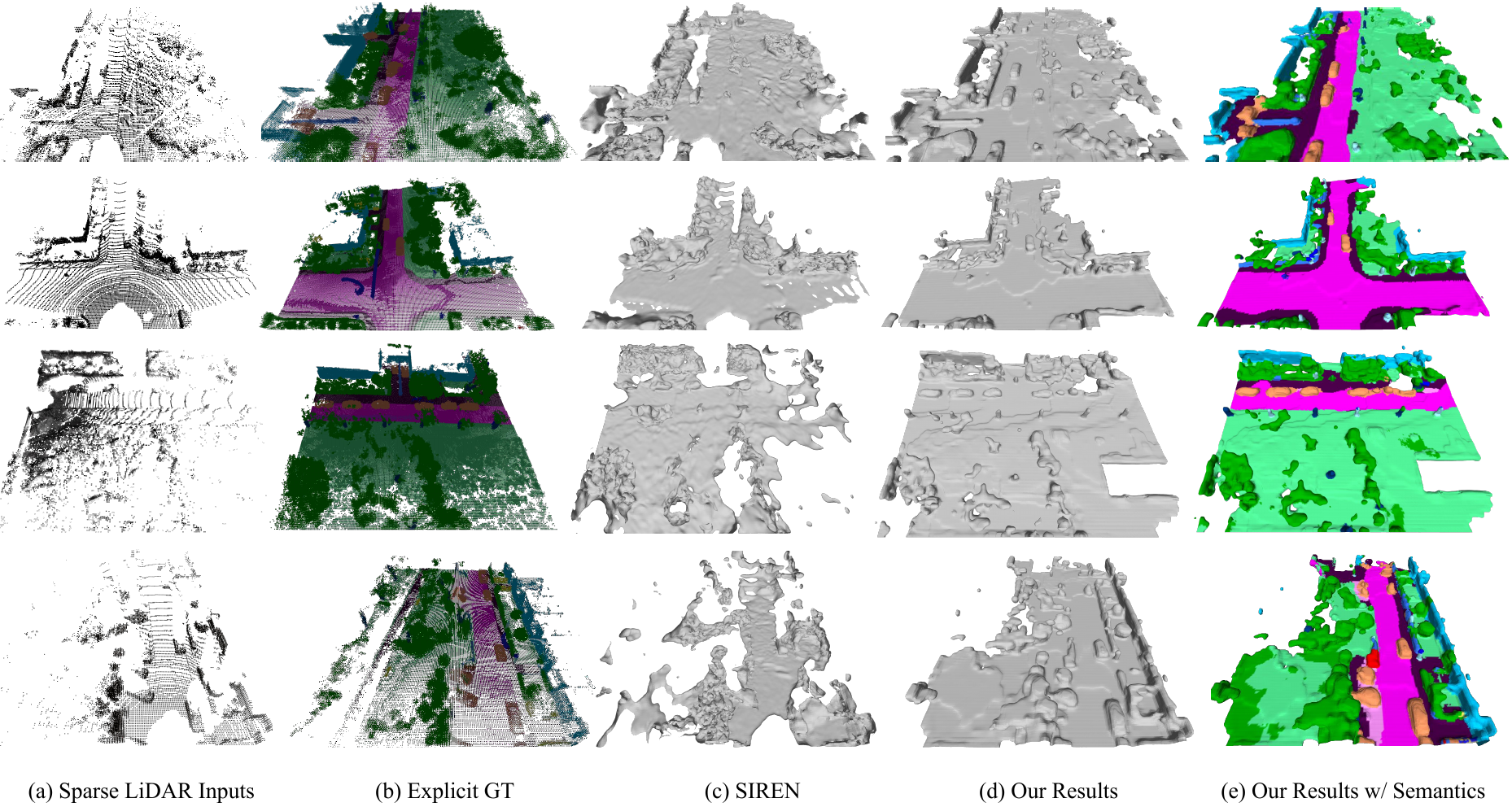}}
\caption{More qualitative results (cont.).}
\label{fig:more_qualitative_1}
\end{figure*}